\documentclass[11pt]{article}

\usepackage[final]{acl}
\usepackage{times}
\usepackage{latexsym}
\usepackage[T1]{fontenc}
\usepackage[utf8]{inputenc}
\usepackage{microtype}
\usepackage{inconsolata}
\usepackage{graphicx}

\usepackage{amsmath}
\usepackage{booktabs}
\usepackage{tabularx}
\usepackage{enumitem}
\usepackage{makecell}

\newcommand{\model}{\textsc{MemProbe}}

\title{Probing Stability–Plasticity Tradeoffs in Agent Memory through Cognitive Experimental Paradigms}

\author{Jiaqi Ding \\
  Department of Computer Science \\
  UNC-Chapel Hill \\
  \texttt{jiaqid@cs.unc.edu} \\\And
  Guorong Wu \\
  Department of Computer Science \\
  Department of Psychiatry \\
  UNC-Chapel Hill \\
  \texttt{guorong\_wu@med.unc.edu} \\}

\begin{document}
\maketitle
\begin{abstract}
Agent memory systems are increasingly used to maintain long-term user preferences, task states and evolving facts, but current evaluations often collapse memory behavior into final-answer accuracy. 
We introduce \model{}, a cognitive-science-inspired framework for diagnosing stability--plasticity tradeoffs in agent memory.
The framework is motivated by a core insight from cognitive memory research: memory is reconstructive and shaped by interference, source reliability, reinforcement, and reactivation.
\model{} turns this insight into four reusable experimental paradigms (interference, misinformation, consolidation strength, and reconsolidation window) that manipulate when a memory should be updated, preserved, or treated as uncertain.
It further decomposes correctness into behavioral profiles that reveal how systems update, preserve, attribute, and temporally organize information. We instantiate these paradigms in a 56-episode diagnostic suite and evaluate six incremental memory systems under a unified protocol. Results show that systems with similar aggregate scores exhibit distinct behavioral profiles.
\model{} provides such a diagnostic lens, turning aggregate performance into interpretable profiles of memory maintenance over time.
Code is available at \url{https://github.com/jq-ding/MemProbe}.
\end{abstract}

\section{Introduction}

\begin{figure}
    \centering
    \includegraphics[width=1\linewidth]{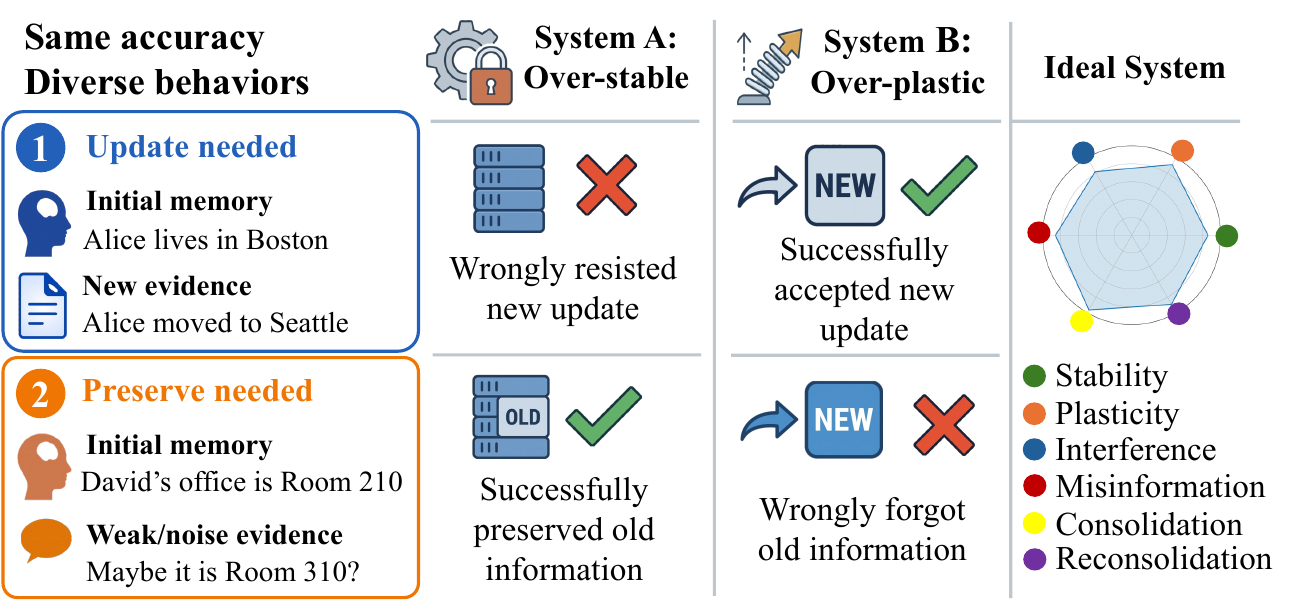}
    \caption{\small{{Illustrative depiction of how the same accuracy can mask distinct memory-update behaviors, motivating behavioral evaluation beyond final-answer correctness.}}}
    \label{fig:intuition}
\end{figure}

Two agent memory systems can achieve identical accuracy but behave very differently when updating memory: one may aggressively overwrite prior memories upon seeing new evidence; another may retain outdated information unless explicitly corrected. Identical scores can thus mask very different risks for long-term memory (Figure~\ref{fig:intuition}). 

This distinction has become increasingly important because modern agent memory systems differ substantially in how they store, retrieve, compress, invalidate, and update information. Retrieval-only and Full-context systems often defer conflict resolution to query time, where a reader model interprets the available evidence. Other systems maintain memory during ingestion by extracting structured facts, compressing interaction histories into evolving summaries, or applying explicit operations such as \textsc{Add}, \textsc{Update}, \textsc{Delete}, and \textsc{Noop}. Therefore, these mechanisms can produce different update tendencies even when their final answers are similar.

Existing benchmarks have substantially expanded the scope of long-term memory evaluation. Long-context and conversational memory benchmarks such as LongMemEval~\cite{wu2024longmemeval}, LoCoMo~\cite{maharana2024evaluating} and MemBench~\cite{Tan2025MemBench} evaluate temporal reasoning, knowledge updates, and multi-session understanding. More recent memory-agent benchmarks such as MemoryAgentBench~\cite{hu2025evaluating}, MemoryStress~\cite{memorystress}, AMA-Bench~\cite{zhao2026amabenchevaluatinglonghorizonmemory}, MemoryArena~\cite{he2026memoryarena} and EverMemBench~\cite{hu2026evaluatinglonghorizonmemorymultiparty} further move toward incremental memory maintenance, contradiction handling, agentic trajectories, and temporally evolving decisions. These benchmarks are valuable and have pushed the field beyond simple recall.

However, their primary outcomes remain task-level correctness measures such as accuracy, F1, recall or task success rate. Such measures tell us whether a system produced the correct final answer, but not what memory-update behavior produced that answer. A system may answer a contradiction case correctly 
because it retrieves the right evidence, because it has learned a reliable update policy, or simply because the test distribution favors update-needed examples. Conversely, a system with high aggregate accuracy may still be over-plastic or over-stable. As agent memory systems become more diverse, evaluating only final-answer correctness risks collapsing distinct update profiles into a single score.

We therefore introduce \model{}, a cognitive-science-inspired framework for behaviorally evaluating stability--plasticity tradeoffs in agent memory systems. Rather than assuming access to an agent's internal memory, \model{} controls the input sequence and observes the resulting behavior.
This follows the experimental logic of cognitive memory research: when internal memory states cannot be directly observed, controlled behavioral paradigms can reveal systematic memory tendencies. 

\model{}'s primary contribution is behavioral measurements and a set of reusable paradigm specifications for diagnosing memory-update dynamics.
We make three contributions:

\textbf{First}, we introduce a paradigm-specification framework for evaluating agent memory update behavior. \model{} defines four diagnostic paradigms: Interference~\citep{underwood1957interference,barnes1959fate}, 
Misinformation~\citep{loftus1975leading}, 
Consolidation Strength~\citep{ebbinghaus1913memory,mcclelland1995there}, 
and Reconsolidation Window~\citep{nader2000fear}. Each paradigm specifies what information is encoded, what perturbation is introduced, what variables are controlled, what probes are asked, and what behavioral signatures are expected. 

\textbf{Second}, we define behavioral metrics that decompose aggregate correctness into multi-dimensional stability--plasticity profiles. 
Based on these metrics, \model{} further generates a systematic behavioral profile for each evaluated memory system, summarizing its strengths, weaknesses, dominant failure modes and stability--plasticity tendencies in a structured report.

\textbf{Third}, we instantiate the paradigms in a compact diagnostic suite and apply them to existing memory systems. Empirically, \model{} shows that high-scoring systems can still differ markedly in how they maintain evolving information, motivating behavioral profiles rather than single-score evaluation. At the same time, \model{} is not limited to this fixed suite: we provide the trial structure, latent specification, controlled variables, probe types, gold-label requirements, and scoring protocol needed for researchers to construct customized suites in their own domains.

The rest of the paper is organized as follows: 
Sec.\ref{sec:paradigm-specification} introduces the \model{} paradigm specification.
Sec.\ref{sec:experimental-setup} describes the evaluated systems, input protocols and behavioral metrics.
Sec.\ref{sec:results-analysis} reports results and analyzes stability--plasticity profiles, paradigm-wise behavior and failure modes.

\section{Related Work}
\label{sec:related-work}

\textbf{Agent Memory Systems.}
Modern agent memory systems~\citep{Park2023GenerativeAgents,simplemem2026, zhong2024memorybank, memgpt, fang2025lightmem, tan-etal-2025-prospect} differ substantially in how they store, retrieve, compress, invalidate, and update information. 
Extractive systems such as Mem0~\citep{mem0} and Zep~\citep{rasmussen2025zeptemporalknowledgegraph} instead maintain structured memories by writing, merging or invalidating facts during ingestion. Compression-based systems like Observational Memory~\citep{barnes2026observationalmemory} update an implicit summary or observation state through repeated summarization. Policy-based systems such as Memory-R1~\cite{memoryr1} and AgeMem~\citep{yu2026agenticmemorylearningunified} learn explicit memory operations. 
These mechanisms can produce different update tendencies even when their final answers are similar, motivating evaluation that looks beyond task accuracy to the underlying memory behavior.

\noindent{}\textbf{Agent Memory Benchmarks.}
Recent benchmarks~\citep{hu2026evaluatinglonghorizonmemorymultiparty, wang2026evomembenchbenchmarkingagentmemory, bian2026realmembenchmarkingllmsrealworld} evaluate how LLMs and agents retain and use information across long interaction histories. LoCoMo~\citep{maharana2024evaluating} and LongMemEval~\citep{wu2024longmemeval} probe long-horizon conversational recall through multi-session dialogue tasks, while MemoryAgentBench~\citep{hu2025evaluating} and MemBench~\citep{Tan2025MemBench} move toward incremental memory maintenance, treating memory as something accumulated through interaction rather than read from a static long context.

Another group of benchmarks stresses memory systems with contradiction, noise, and evolving facts. MemoryStress simulates 1{,}000 sessions with contradictions, fading memories, and accumulated noise~\citep{memorystress}, and MemoryArena evaluates interdependent multi-session tasks where later success depends on distilling earlier actions and feedback~\citep{he2026memoryarena}. Concurrent with our work, MemConflict~\citep{tao2026memconflict} examines whether long-term memory systems retrieve and apply memories that are temporally valid, factually correct, and contextually applicable under conflicting evidence. \model{} is complementary, rather than scoring task accuracy under long horizons or conflict, it treats conflict as one manifestation of a broader stability--plasticity trade-off and uses cognitive-science-inspired paradigms to diagnose memory update behaviors.

\begin{figure*}[ht]
    \centering
    \includegraphics[width=0.95\linewidth]{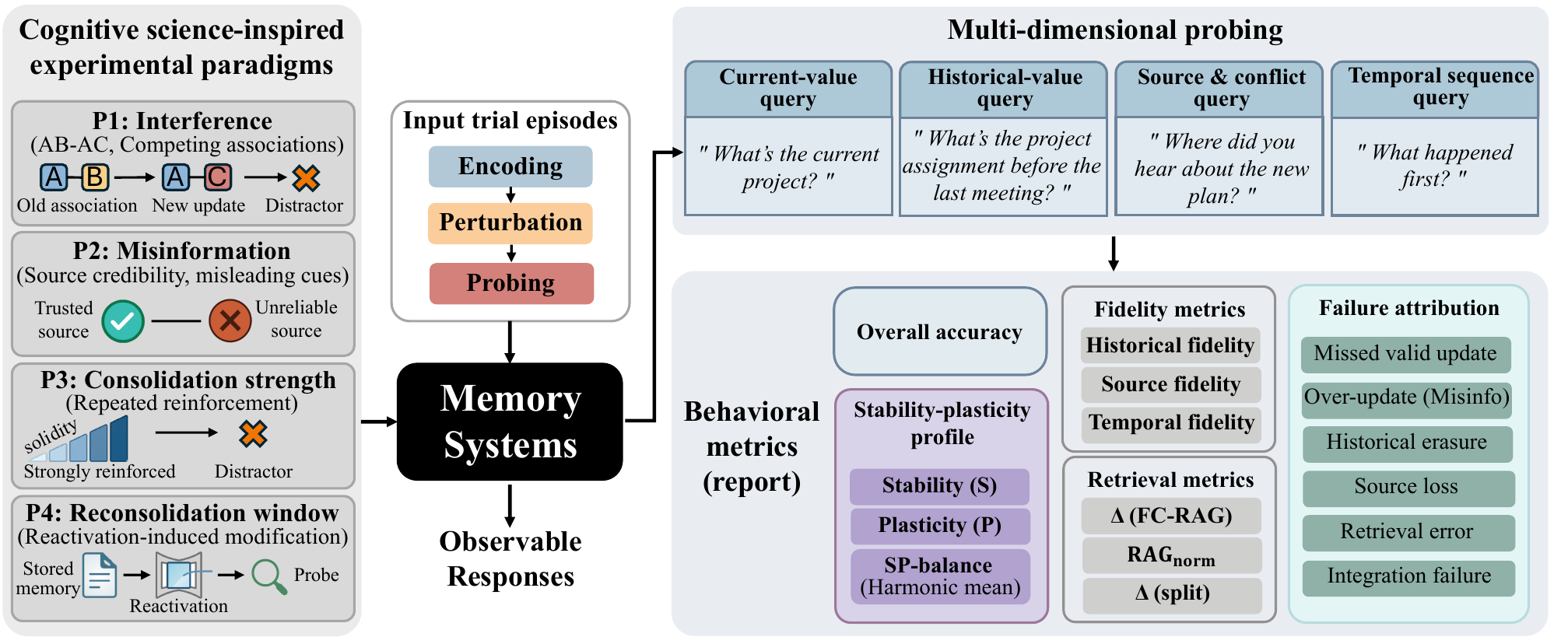}
    \caption{\small{
    Overview of the \model{} evaluation framework. Cognitive-science-inspired paradigms generate controlled trial episodes that are processed by a memory system and queried through multi-dimensional probes. Observable responses are converted into behavioral stability--plasticity profiles.
    }}
    \label{fig:framework}
\end{figure*}

\section{From Cognitive Memory Paradigms to Agent Memory Diagnosis}
\label{sec:paradigm-specification}

\model{} formulates evaluation as a behavioral diagnosis problem: controlled contexts manipulate memory conditions, and targeted probes reveal patterns of update behaviors.

\subsection{Diagnostic Paradigms from Cognitive Memory Research}

Many agent memory systems differ in how they store, retrieve, compress or update information. Therefore, following the logic of cognitive memory experiments, \model{} diagnoses memory behavior through observable responses rather than implementation-specific memory states (Figure~\ref{fig:framework}).

\paragraph{Paradigm I: Interference.}
This paradigm is motivated by classic $AB-AC$ paired-associate learning studies~\cite{underwood1957interference,barnes1959fate}. Participants first learn an association $A$--$B$ and later learn a competing association $A$--$C$. When probed with $A$, producing the old value $B$ after learning $C$ reflects proactive interference, whereas producing $C$ when asked for the earlier association reflects retroactive interference. The paradigm therefore separates successful updating from preservation of historical memory.

In agent memory, the same logic captures competition among an established memory, a legitimate update and a later distractor. 
This paradigm probes excessive stability, where valid updates fail, and excessive plasticity, where distractors intrude as new memories.

\paragraph{Paradigm II: Misinformation.}
This paradigm studies how post-event misleading information can distort memory for an original event~\cite{loftus1974reconstruction,loftus1975leading}. Its key insight is that later information affects memory not only through its content, but also through its source and credibility: misleading details from trusted sources are more likely to be incorporated, whereas unreliable sources are more likely to be discounted.

In agent memory, conflicting values may come from users, assistants, tools, retrieved documents, stale trackers or third-party statements. The central question is whether the system distinguishes evidence from mere mention. A reliable user correction or trusted tool update may warrant revision, while a hallucinated assistant statement, rumor or non-target value should not overwrite memory. This paradigm tests source-sensitive updating and misinformation-driven over-updating.

\paragraph{Paradigm III: Consolidation Strength.}
This paradigm reflects that strongly reinforced memories are more resistant to disruption. Overlearning improves retention~\cite{ebbinghaus1913memory}, and interference studies suggest that repeatedly learned associations are harder to overwrite than weakly learned ones~\cite{underwood1957interference}. Related reconsolidation work further shows that more consolidated memories can be less susceptible to modification even after reactivation~\cite{milekic2002temporally}.

For agent memory, this means update behavior should depend on prior support. A fact repeatedly confirmed across interactions should be harder to overwrite than a one-off statement, especially under weak or unreliable challenges. However, strong reliable evidence should still allow revision. This paradigm tests whether a system treats all memories as equally plastic or adjusts stability and plasticity according to memory strength.

\paragraph{Paradigm IV: Reconsolidation Window.}
This paradigm shows that recalling a consolidated memory can temporarily make it modifiable~\cite{nader2000fear,schiller2010preventing}. After reactivation, new information may update or distort the original memory before it stabilizes again. This effect can also be selective: directly reactivated memories are more susceptible to modification than indirectly activated ones~\cite{dekebiec2006directly}.

In agent memory, an old memory may be explicitly recalled before new evidence appears. Reactivation can help compare old and new information, but it should not make the system indiscriminately accept weak speculation or misinformation. This paradigm tests whether reactivation changes update sensitivity, and whether that sensitivity is selective to reliable evidence rather than causing reactivation-induced over-updating.

\subsection{Controlled Trial Design}

Each \model{} trial is an ordered episode in the \textit{encoding--perturbation--probing} structure designed to expose a specific memory-update behavior. 
In an \textit{encoding} stage, an initial target memory is established. It may then include filler interactions, near-miss facts or distractors to create a realistic memory context and prevent the task from reducing to simple recency matching.

The central manipulation occurs in the \textit{perturbation} stage. Depending on the paradigm, the perturbation may introduce a valid update, a conflicting but unreliable value, a weak challenge to an established memory or an explicit reactivation of a prior memory. Some trials also include post-perturbation context, which tests whether the system is overly sensitive to surface similarity, later mentions or irrelevant recency cues.

Finally, the episode ends with targeted \textit{probes}. These probes query the system about the current state, previous state, source of information, conflict status or temporal update sequence. 
What must be preserved is the controlled relation among the initial memory, the perturbation, the expected behavior and the probe.
App.~\ref{app:suite_details} provides the construction details of our demonstration suite.

\section{\model{} Evaluation Protocol}
\label{sec:experimental-setup}

We now specify how \model{} trials are evaluated in practice. The goal is not to collapse systems into a single ranking, but to make their responses comparable through a shared input protocol and a common set of behavioral measurements. 

\subsection{Systems and Input Protocol}

\begin{figure*}
    \centering
    \includegraphics[width=0.92\linewidth]{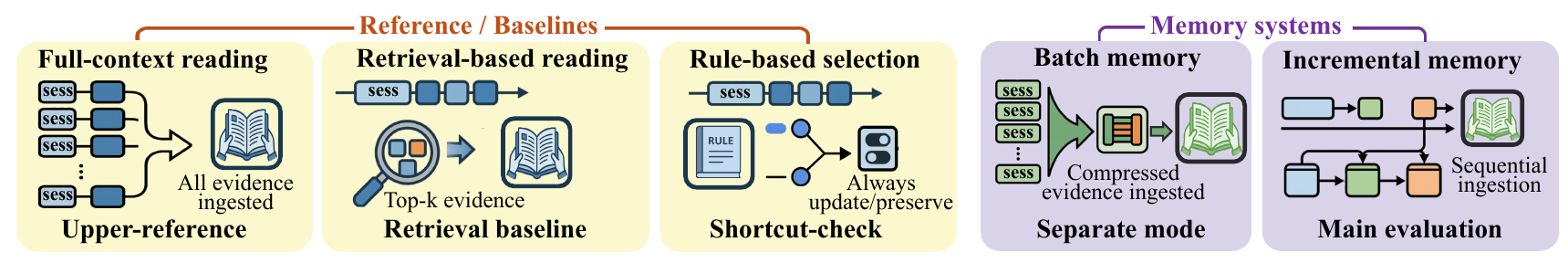}
    \caption{Input protocols used in \model{}. Full-context reading, retrieval-based reading, heuristic baselines, batch memory, and incremental memory correspond to different computational interpretations of memory and are therefore reported separately. Incremental memory is the primary setting for evaluating stability--plasticity behavior.}
    \label{fig:evaluation_modes}
\end{figure*}

\model{} supports several input protocols (Figure~\ref{fig:evaluation_modes}). 
The first set of input protocols are used as references/baselines (yellow boxes in Figure~\ref{fig:evaluation_modes}). A \textbf{Full-context reader} receives the complete episode at probe time and serves as an upper-reference for whether the trial is answerable under complete evidence. Retrieval-based baselines~\cite{RAG}, including \textbf{Naive RAG} (retrieves the top-$k$ sessions independently for each probe), \textbf{Time-aware RAG} (additionally incorporates session order and temporal metadata) and \textbf{Oracle RAG} (provides gold-relevant evidence sessions to distinguish retrieval failure from reader reasoning failure), index sessions as retrievable units and answer each probe from retrieved evidence. We also include \textbf{heuristic baselines} that use simple rule-based candidate selection, such as preferring the most recent value or selecting values from specific source types. These heuristics are not intended as memory systems, but serve as sanity checks for shortcut vulnerability. App.~\ref{app:suite_validation} reports schema checks, probe-level validation, heuristic baselines, and cross-model validation results.

Our primary evaluation setting is \textbf{incremental memory maintenance} (purple boxes in Figure~\ref{fig:evaluation_modes}) because it requires the system to decide during ingestion whether new evidence should update, preserve or coexist with prior memory. For each episode, the memory store is reset and sessions are provided in chronological order. At probe time, the system must answer using its maintained memory state, retrieval interface or internal memory mechanism. Formally, for an episode $e$ with sessions $\{s_t\}_{t=1}^T$ and probes $\{q_j\}_{j=1}^m$, the protocol is:

\begin{equation}
\begin{aligned}
M_0      &\leftarrow \mathrm{Reset}(),\\
M_t      &\leftarrow \mathrm{Ingest}(M_{t-1}, s_t),
          && t=1,\ldots,T,\\
\hat{a}_j &\leftarrow \mathrm{Query}(M_T, q_j),
          && j=1,\ldots,m.
\end{aligned}
\end{equation}
We evaluate six incremental memory systems under the same sequential-ingestion protocol: \textbf{Mem0}~\cite{mem0}, \textbf{Zep/Graphiti}~\cite{rasmussen2025zeptemporalknowledgegraph}, \textbf{LangMem}~\cite{langchain2025langmem}, \textbf{Cognee}~\cite{cogneemarkovic2025optimizinginterfaceknowledgegraphs}, \textbf{A-MEM}~\cite{xu2026mem}, and \textbf{MemoryOS}~\cite{kang2025memoryosaiagent}. These systems cover extraction-based memory, temporal knowledge-graph memory, framework-level memory primitives, graph-vector hybrid memory, agentic self-organizing memory and hierarchical memory-OS designs. For consistency, all systems are fed the same ordered sessions and are probed using a unified reader over their retrieved or extracted memory. When a system requires an internal LLM for memory extraction or consolidation, we use the same internal model whenever the interface permits.

\subsection{Behavioral Metrics}

All metrics are computed from structured probe scores. 
A probe score records whether the system's answer matches the structured gold fields required by that probe, such as the current/previous values.
We define the full probe taxonomy and scoring procedure in App.~\ref{app:probe_scoring}.
Let $\mathcal{Q}$ denote the set of all probes and let $c(q,S)$ be an indicator of whether system $S$ answers probe $q$ correctly under this scoring procedure. We define accuracy over any probe subset $\mathcal{D}$ as:
\begin{equation}
A(S;\mathcal{D})
=
\frac{1}{|\mathcal{D}|}
\sum_{q \in \mathcal{D}} c(q,S).
\label{eq:subset_accuracy}
\end{equation}
Overall accuracy is computed by applying Eq.~\eqref{eq:subset_accuracy} to all probes:
$
\mathrm{Acc}(S)=A(S;\mathcal{Q}).
$

Unless noted otherwise, confidence intervals are episode-level bootstrap intervals (2{,}000 resamples, fixed seed), and system comparisons use paired bootstrap on the same resampled episodes.

\paragraph{Stability--plasticity metrics.}
Let $\mathcal{U}$ denote probes from trials where the expected behavior is to accept a valid update, and let $\mathcal{P}$ denote probes from trials where the expected behavior is to preserve the existing memory. Using Eq.~\eqref{eq:subset_accuracy}, we define plasticity and stability as:
\begin{equation}
\begin{aligned}
\mathrm{Plasticity}(S) &= A(S;\mathcal{U}), \\
\mathrm{Stability}(S) &= A(S;\mathcal{P}).
\end{aligned}
\label{eq:plasticity_stability}
\end{equation}
The complementary condition-level error rates are:
\begin{equation}
\begin{aligned}
\mathrm{UpdateError}(S) &= 1 - \mathrm{Plasticity}(S), \\
\mathrm{PreserveError}(S) &= 1 - \mathrm{Stability}(S).
\end{aligned}
\label{eq:update_errors}
\end{equation}
To provide a compact summary across the two conditions, we report their
harmonic mean:
\begin{equation}
\mathrm{SP\text{-}Balance}(S)
=
\frac{
2\,\mathrm{Plasticity}(S)\,\mathrm{Stability}(S)
}{
\mathrm{Plasticity}(S)+\mathrm{Stability}(S)
}.
\label{eq:sp_balance}
\end{equation}
Eq.~\eqref{eq:sp_balance} penalizes systems that perform well on only one side of the stability--plasticity tradeoff.

\paragraph{Historical, source and temporal fidelity.}
We derive fidelity metrics from the corresponding probe types defined in App.~\ref{app:probe_types}. Historical fidelity is measured by previous-value accuracy, source fidelity by source and conflict-status accuracy, and temporal fidelity by temporal-probe accuracy. These metrics measure whether a system preserves not only the current answer, but also the prior state, provenance and update sequence behind that answer.

\begin{table*}[ht]
\centering
\small
\resizebox{\linewidth}{!}{
\setlength{\tabcolsep}{3pt}
\begin{tabular}{lcccccccccc}
\toprule
\textbf{Metric}
& \makecell{\textbf{Oracle RAG}\\ \textbf{(diag.)}}
& \textbf{Full-context}
& \textbf{Mem0}
& \textbf{Graphiti}
& \textbf{LangMem}
& \textbf{Cognee}
& \textbf{A-MEM}
& \textbf{MemoryOS}
& \makecell{\textbf{Naive }\\ \textbf{RAG}}
& \makecell{\textbf{Time-aware }\\ \textbf{RAG}} \\
\midrule
\makecell{\textbf{Overall} \\ \textbf{[95\% CI]}}
& \makecell{93.3 \\ {[89.7, 96.4]}}
& \makecell{88.8 \\ {[84.4, 92.9]}}
& \makecell{41.5 \\ {[35.7, 46.9]}}
& \makecell{62.5 \\ {[55.8, 68.8]}}
& \makecell{65.2 \\ {[57.6, 72.3]}}
& \makecell{81.7 \\ {[76.3, 86.6]}}
& \makecell{88.4 \\ {[83.9, 92.4]}}
& \makecell{60.3 \\ {[53.6, 66.5]}}
& \makecell{68.3 \\ {[60.3, 76.3]}}
& \makecell{68.8 \\ {[61.2, 75.9]}} \\
\bottomrule
\end{tabular}}
\caption{
Overall behavioral performance across references, incremental memory systems and retrieval baselines.
}
\label{tab:overall_memory_performance}
\end{table*}

\paragraph{Retrieval and evidence-integration metrics.}
For retrieval-based diagnostic baselines, we compare retrieved-context performance against full-context reading. Let $\mathrm{Acc}_{\mathrm{FC}}$ denote full-context accuracy and let $\mathrm{Acc}_{r}$ denote the accuracy of a retrieval-based condition $r$. We define
$
\Delta_{\mathrm{FC}}(r)
=
\mathrm{Acc}_{\mathrm{FC}}
-
\mathrm{Acc}_{r},
$
and
$
\mathrm{RAG}_{\mathrm{norm}}(r)
=
\frac{\mathrm{Acc}_{r}}
{\mathrm{Acc}_{\mathrm{FC}}}.
$
Here, $\Delta_{\mathrm{FC}}$ measures the performance loss relative to full evidence, while $\mathrm{RAG}_{\mathrm{norm}}$ measures the fraction of full-context performance recovered by retrieved evidence. Values above 1 are possible for diagnostic evidence-selection baselines such as Oracle RAG, because they remove filler and near-miss sessions and provide only clean evidence.

For split-perturbation episodes, where relevant evidence is distributed across sessions, we define
$
\Delta_{\mathrm{split}}
=
\mathrm{Acc}_{\mathrm{split}}
-
\mathrm{Acc}_{\mathrm{non\text{-}split}}.
$
Negative values therefore indicate a split penalty, while positive values indicate better performance on split episodes.

\paragraph{Failure attribution.}
Incorrect predictions are assigned to diagnostic failure categories based on the structured scores described in App.~\ref{app:scoring_procedure}. Retrieval-based systems may fail because relevant evidence is missing, only part of a split perturbation is retrieved, or the reader misinterprets retrieved evidence. Incremental systems may fail by missing valid updates, over-updating on weak evidence, overwriting historical values, losing source information, or retrieving the wrong memory at query time. These categories are reported separately from aggregate accuracy.

\section{Results and Analysis}
\label{sec:results-analysis}

All incremental memory systems are evaluated with a shared protocol: each episode uses a fresh memory store, sessions are ingested sequentially, retrieved/extracted memory is passed to a unified Gemini-3-Flash reader, and answers are scored using the same structured JSON scorer. \S\ref{sec:reader-ablation} verifies that the resulting profiles do not depend on which reader is used.
For systems requiring an internal LLM for extraction, we use Gemini-2.5-Flash when configurable. The LLM-selection rationale and full implementation details are provided in App.~\ref{app:implementation}.

\subsection{Overall Memory-System Performance}
\label{sec:overall-memory-performance}

Table~\ref{tab:overall_memory_performance} provides a coarse entry point into system behavior. We report overall accuracy to summarize performance, but do not treat it as the final evaluation criterion. We separate diagnostic references from real memory systems. Among incremental memory systems, A-MEM performs best, reaching 88.4\% overall accuracy and nearly matching the Full-context upper reference at 88.8\%. Cognee ranks second with 81.7\% overall accuracy, followed by LangMem, Graphiti, and MemoryOS in the middle range.

\begin{figure}[ht]
    \centering
    \includegraphics[width=1\linewidth]{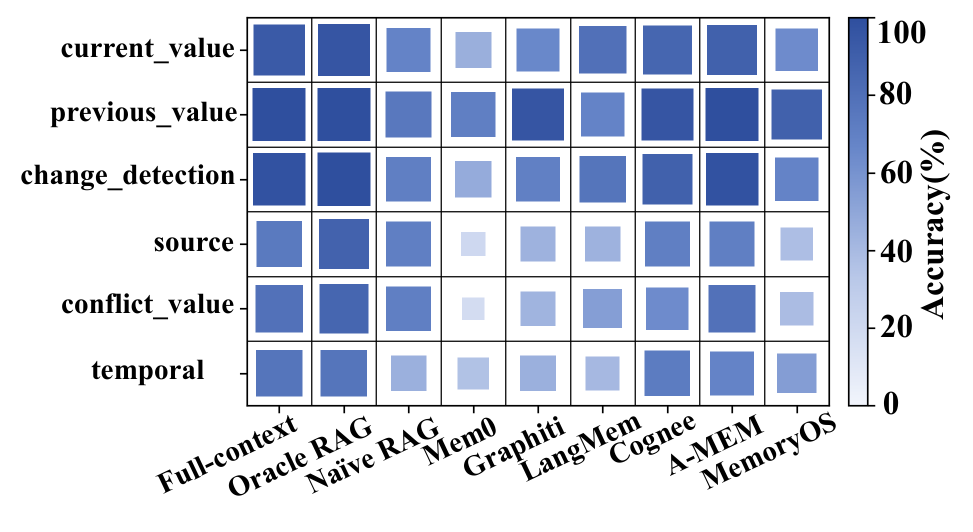}
    \caption{
Probe-level performance across systems. 
Many systems perform well on current-value, previous-value and change-detection probes, but degrade on source, conflict-value and temporal probes.
}
    \label{fig:probe}
\end{figure}

Figure~\ref{fig:probe} further decomposes performance by probe type (Probe definitions are provided in App.~\ref{app:probe_types}). Most systems perform better on current-value, previous-value, and change-detection probes than on source, conflict-value, and temporal probes. 
This suggests that many memory systems can retain or retrieve factual values, but often fail to preserve provenance and update-sequence information. 

These aggregate scores provide only a coarse summary. As the following sections show, systems with similar overall accuracy can differ sharply in their update tendencies, historical retention, provenance preservation, temporal reasoning, and robustness to split evidence. We therefore decompose the results into stability--plasticity profiles, paradigm-wise behavior, and failure modes below.

\subsection{Stability--Plasticity Profiles}
\label{sec:stability-plasticity-results}

Aggregate accuracy alone cannot distinguish systems that update too aggressively from systems that preserve outdated memories too strongly. We therefore compare systems by plasticity, stability and the corresponding condition-level error rates. 
Plasticity measures whether a system accepts valid updates, while stability measures whether it resists weak or unreliable evidence.

\begin{table}[ht]
\centering
\small
\resizebox{\linewidth}{!}{
\setlength{\tabcolsep}{3pt}
\begin{tabular}{lrrrrr}
\toprule
\textbf{System} & \textbf{Plast.} & \textbf{Stab.} & \textbf{SP-Bal} & \textbf{PreserveErr.} & \textbf{UpdateErr.} \\
\midrule
Mem0           & 28.6 & 54.5 & 37.5 & 45.5 & 71.4 \\
Graphiti       & 58.9 & 66.1 & 62.3 & 33.9 & 41.1 \\
LangMem        & 63.4 & 67.0 & 65.1 & 33.0 & 36.6 \\
Cognee         & 75.0 & 88.4 & 81.1 & 11.6 & 25.0 \\
A-MEM          & 91.1 & 85.7 & 88.3 & 14.3 & 8.9 \\
MemoryOS       & 54.5 & 66.1 & 59.7 & 33.9 & 45.5 \\
\bottomrule
\end{tabular}}
\caption{
Stability--plasticity metrics for incremental memory systems.
}
\label{tab:stability_plasticity_results}
\end{table}

\begin{figure}[ht]
\centering
\includegraphics[width=0.8\linewidth]{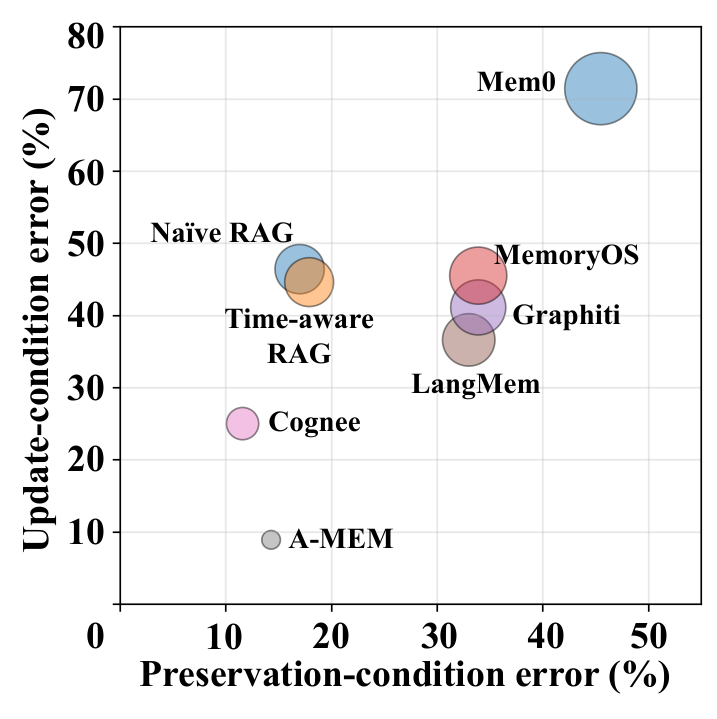}
\caption{
Stability--plasticity plane. Systems closer to the lower-left corner have better SP-balance.
}
\label{fig:sp_plane}
\end{figure}

\begin{table*}[t]
\centering
\small
\begin{tabularx}{\linewidth}{l *{5}{>{\centering\arraybackslash}X}}
\toprule
\textbf{System} & \textbf{Interference} & \textbf{Misinformation} & \textbf{Consolidation} & \textbf{Reconsolidation} & \textbf{Overall} \\
\midrule
Oracle RAG (diag.) & 89.3 & 92.9 & 98.2 & 92.9 & 93.3 \\
Full-context       & 82.1 & 91.1 & 91.1 & 91.1 & 88.8 \\
\midrule
Mem0               & 41.1 & 46.4 & 33.9 & 44.6 & 41.5 \\
Graphiti           & 53.6 & 60.7 & 75.0 & 60.7 & 62.5 \\
LangMem            & 60.7 & 66.1 & 69.6 & 64.3 & 65.2 \\
Cognee             & 80.4 & 80.4 & 85.7 & 80.4 & 81.7 \\
A-MEM              & 82.1 & 85.7 & 96.4 & 89.3 & 88.4 \\
MemoryOS           & 66.1 & 55.4 & 57.1 & 62.5 & 60.3 \\
\midrule
Naive RAG          & 67.9 & 75.0 & 60.7 & 69.6 & 68.3 \\
Time-aware RAG     & 67.9 & 73.2 & 66.1 & 67.9 & 68.8 \\
\bottomrule
\end{tabularx}
\caption{
Paradigm-wise accuracy across the four paradigms. Interference is difficult even for strong references, while consolidation sharply separates structured memory systems from flatter extraction or retrieval systems.
}
\label{tab:paradigm_results}
\end{table*}

Table~\ref{tab:stability_plasticity_results} reports the stability--plasticity metrics. A-MEM has the best balance among memory systems, with 91.1\% plasticity, 85.7\% stability, and an SP-Balance of 88.3\%. Cognee is more stability-biased: it achieves the highest stability at 88.4\%, but lower plasticity at 75.0\%. LangMem and Graphiti occupy a middle region with different behaviors: LangMem consolidates aggressively and loses history, while Graphiti retains historical values but struggles to update and preserve source information. 

Figure~\ref{fig:sp_plane} plots systems in the stability--plasticity plane. The ideal region is the lower-left corner, where errors are low under both conditions. 
A-MEM is closest to this region. Cognee is also strong but more conservative, with fewer errors when preservation is required and more when an update is required. 
Mem0 lies in the least desirable region. 

The three mid-range systems (Graphiti, LangMem, MemoryOS) are statistically indistinguishable in overall accuracy (all pairwise $p>0.37$), and thus directly fall into the ``same accuracy'' condition. But their behavior differs once correctness is decomposed by probe type. Graphiti preserves historical values far better than LangMem (previous-value $96.4$ vs. $67.9$, 95\% CI $[+9.1, +48.3]$, $p=0.002$), a gap that survives Bonferroni correction across all 18 system--probe comparisons, while the
current-value gap runs in the opposite direction ($66.1$ vs. $80.4$, $\Delta=-14.3$pp). Two systems with the same aggregate score therefore realize opposite maintenance strategies: one biased toward retaining the past, the other toward tracking the present.

These results illustrate the central motivation of \model{}: final accuracy alone cannot reveal whether a memory system is over-plastic, over-stable or balanced.

\subsection{Paradigm-wise Analysis}
\label{sec:paradigm-wise-results}

We next analyze system behavior across the four \model{} paradigms. Each paradigm contains 14 episodes and 56 scored probes. Table~\ref{tab:paradigm_results} reports paradigm-wise accuracy.

Several patterns emerge. \textbf{First}, interference is the hardest paradigm for the strongest references: Full-context reaches 82.1\%, A-MEM 82.1\%, and Oracle RAG 89.3\%. This suggests that competing similar facts stress memory behavior even when clean evidence is available. \textbf{Second}, consolidation sharply separates systems. A-MEM reaches 96.4\%, and Graphiti achieves its best paradigm score at 75.0\%, suggesting that note-based and graph-based structures help retain accumulated evidence. 
Mem0 performs less well on consolidation in our setting, suggesting that repeated reinforcement is not always preserved by flat extracted-memory representations.
\textbf{Third}, MemoryOS shows a different pattern from most other systems: it performs best on interference but worst on misinformation and consolidation. This suggests that its heat-promotion mechanism may keep recent or competing items accessible while failing to preserve reinforced history. 
\textbf{Finally}, misinformation favors systems that preserve raw or well-attributed evidence. Retrieval-style access performs relatively well, A-MEM and Cognee remain strong.
In contrast, systems that aggressively extract, merge or rewrite memories often blur provenance during memory construction. App.~\ref{app:correlation} further reports how much independent signal each decomposition axis carries.

\subsection{Reader Ablation}
\label{sec:reader-ablation}

To check whether the profiles reflect the memory representation rather than the reader's ability to interpret it, we re-ran all six memory systems on the same stored memories with a second reader, Qwen3-32B, drawn from a different model family than both the main reader (Gemini) and the generator (GPT) used for dialogue surface form. So any difference is attributable to the reader alone.

\begin{table}[ht]
\centering
\small
\setlength{\tabcolsep}{6pt}
\begin{tabular}{lrr}
\toprule
\textbf{System} & \textbf{Gemini} & \textbf{Qwen3} \\
\midrule
A-MEM    & 88.4 & 72.8 \\
Cognee   & 81.7 & 68.3 \\
LangMem  & 65.2 & 60.3 \\
Graphiti & 62.5 & 61.6 \\
MemoryOS & 60.3 & 51.8 \\
Mem0     & 41.5 & 41.5 \\
\bottomrule
\end{tabular}
\caption{Reader ablation. The same stored memories are answered by two readers from different model families. Absolute scores drop under the weaker reader, but the ordering is largely preserved.}
\label{tab:reader_ablation}
\end{table}

Absolute scores drop under Qwen3-32B, which is expected given its weaker source and conflict-status accuracy (Table~\ref{tab:reader_ablation}). However, the ordering is essentially unchanged (Spearman $\rho=0.94$, Kendall $\tau=0.87$). The single exception is LangMem and Graphiti, which swap positions, whose aggregate scores are statistically indistinguishable (\S\ref{sec:stability-plasticity-results}), and whose order is therefore not expected to be stable under any perturbation.

A reader with weaker provenance reasoning lowers every system's score, but it does not change which system retains more provenance than another. The behavioral profiles reported here are therefore properties of the maintained memory rather than artifacts of the selected reader. Full cross-reader results are given in App.~\ref{app:cross-model-validation}.

\subsection{Failure Taxonomy}
\label{sec:failure-taxonomy}

\model{} is designed to identify not only whether a system fails, but how it fails. We therefore group errors into several recurring failure modes: provenance loss, temporal reconstruction failure, distributed-evidence retrieval miss, historical overwrite, under-update, over-update, and extraction fragility.

\begin{table}[ht]
\centering
\small
\setlength{\tabcolsep}{9pt}
\begin{tabular}{lrrrrrrrr}
\toprule
\textbf{System} & \textbf{Hist.} & \textbf{Source} & \textbf{Temp.}  \\
\midrule
Oracle RAG (diag.) & 100.0 & 75.8 & 77.0 \\
Full-context       & 100.0 & 66.1 & 77.0  \\
\midrule
Mem0               & 71.0  & 17.7 & 36.0 \\
Graphiti (Zep)     & 96.0  & 41.9 & 45.0 \\
LangMem            & 68.0  & 46.8 & 41.0  \\
Cognee             & 96.0  & 64.5 & 73.0  \\
A-MEM              & 100.0 & 64.5 & 68.0  \\
MemoryOS           & 89.0  & 33.9 & 55.0  \\
\midrule
Naive RAG          & 75.0  & 64.5 & 45.0  \\
Time-aware RAG     & 86.0  & 59.7 & 45.0 \\
\bottomrule
\end{tabular}
\caption{
Historical, source and temporal fidelity across systems
}
\label{tab:failure}
\end{table}

The sharpest systematic failure is provenance loss. This indicates that these memory systems can store values but often fail to retain who introduced them, whether they were accepted and whether they applied to the target fact. This could in principle reflect a reader that fails to
recover provenance rather than a memory that fails to retain it.
Two observations separate the two. First, as shown in Table~\ref{tab:failure}, Full-context uses the
same reader on complete evidence and reaches 66.1\%, so part of
the difficulty is inherent to source recovery. But systems fall far
below this same-reader ceiling (Graphiti 41.9\%, Mem0 17.7\%), and a gap of this size under an identical reader cannot be a read-out artifact. Second, we manually inspected every source error and labeled it as a storage-side loss (provenance absent from the retrieved memory,
unrecoverable by any reader) or a read-out failure (provenance
present but not recovered). As shown in Table~\ref{tab:error_audit}, read-out errors are roughly constant across systems (6--9), so the
reader does not explain the large differences in source fidelity.
What varies is storage-side loss, which grows as systems retain less
of the original context.
\begin{table}[ht]
\centering
\small
\setlength{\tabcolsep}{6pt}
\begin{tabular}{lrr}
\toprule
\textbf{System} & \textbf{Read-out} & \textbf{Storage-side} \\
\midrule
A-MEM    & 6 & 2 \\
Cognee   & 9 & 1 \\
LangMem  & 9 & 10 \\
Graphiti & 8 & 11 \\
MemoryOS & 9 & 13 \\
Mem0     & 7 & 20 \\
\bottomrule
\end{tabular}
\caption{Manual classification of source-probe errors. Systems are
ordered by how much of the original context they preserve for the
reader. Read-out errors are roughly constant, whereas storage-side
losses grow as systems compress memory more aggressively.}
\label{tab:error_audit}
\end{table}
Temporal fidelity is also weak: most memory systems fall below the Full-context temporal score of 77\%, suggesting that they do not reliably reconstruct update order, trigger and reason.

Another clear failure mode appears in the split-perturbation setting, where the update trigger/source and the accepted current value are placed in separate sessions. Solving such episodes requires retrieving and integrating both perturbation parts, rather than finding a single decisive update sentence.
In Table~\ref{tab:rag_diagnostic_ablation}, Time-aware RAG adds true temporal position labels to retrieved sessions, but improves only marginally over Naive RAG: 68.8\% overall versus 68.3\%, and 33.3\% split accuracy versus 29.2\%. Split current-value (Split CV) and change-detection (Split CD) accuracy remain stuck at 8.3\% for both Naive and Time-aware RAG. This shows that the problem is not missing temporal order information.

In contrast, Oracle RAG reaches 93.3\% overall and 89.6\% on split episodes. Its split current-value accuracy is 91.7\%, and its split change-detection accuracy is 100\%. Since Oracle RAG uses the same reader but receives clean non-filler evidence, the gap between Naive RAG and Oracle RAG isolates retrieval failure as the main bottleneck. In other words, split failures occur because the retriever often fails to recover both perturbation parts, not because the reader cannot reason over the evidence.

\begin{table}[ht]
\centering
\small
\resizebox{\linewidth}{!}{
\setlength{\tabcolsep}{3.5pt}
\begin{tabular}{lcccc}
\toprule
\textbf{Metric}
& \makecell{\textbf{Naive}\\ \textbf{RAG}}
& \makecell{\textbf{Time-aware}\\ \textbf{RAG}}
& \makecell{\textbf{Oracle}\\ \textbf{RAG}}
& \textbf{Full-context} \\
\midrule
${\Delta_{\mathrm{FC}}}$
& 20.5 & 20.0 & -4.5 & -- \\
${\mathrm{RAG}_{\mathrm{norm}}}$
& 0.769 & 0.775 & 1.051 & -- \\
{Non-split}
& 79.0 & 78.4 & 94.3 & 90.9 \\
{Split}
& 29.2 & 33.3 & 89.6 & 81.2 \\
{Split CV}
& 8.3 & 8.3 & 91.7 & 66.7 \\
{Split CD}
& 8.3 & 8.3 & 100.0 & 91.7 \\
\bottomrule
\end{tabular}}
\caption{
Diagnostic RAG ablation for split-perturbation episodes. 
$\Delta_{\mathrm{FC}}$ measures the loss relative to full-context reading, and $\mathrm{RAG}_{\mathrm{norm}}$ measures the fraction of full-context performance recovered by retrieved evidence. Split CV denotes current-value accuracy, and Split CD denotes change-detection accuracy.
}
\label{tab:rag_diagnostic_ablation}
\end{table}

\begin{figure}[ht]
\centering
\includegraphics[width=1\linewidth]{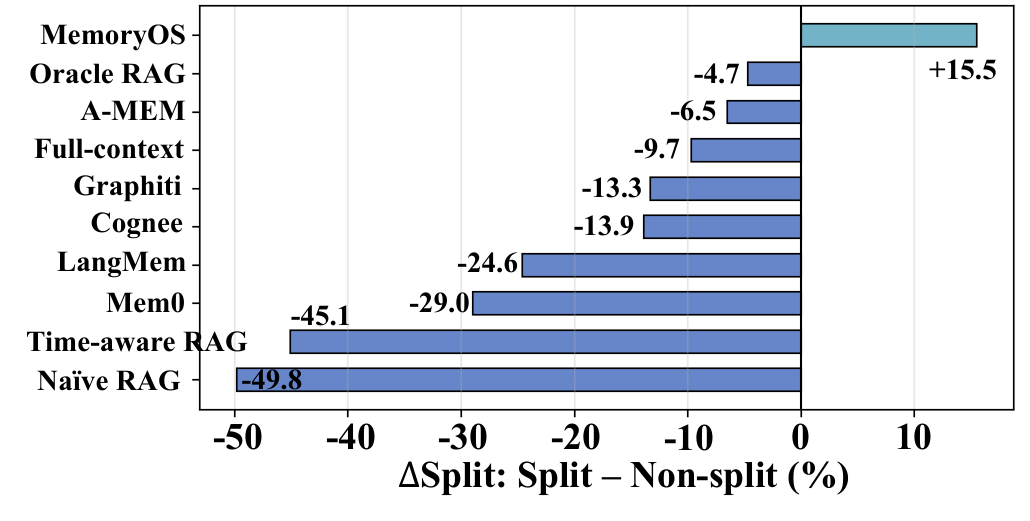}
\caption{
Split penalty/gain by systems. Negative values indicate that performance drops when perturbation evidence is distributed across sessions. 
}
\label{fig:split_delta}
\end{figure}

Finally, Figure~\ref{fig:split_delta} summarizes split penalties across systems. Most systems degrade when evidence is distributed across sessions. MemoryOS is the only system with a positive split delta, but this should be interpreted cautiously because the split subset contains fewer source and conflict probes, which are its weakest categories. Overall, the split results show that distributed evidence integration remains a major weakness for retrieval-based and incremental memory systems.

\vspace{-1em}
\section{Discussion and Conclusion}
\label{sec:discussion-conclusion}

\model{} provides a cognitive-style diagnostic framework for evaluating agent memory systems beyond aggregate final-answer accuracy. By organizing memory evaluation around interference, misinformation, consolidation, and reconsolidation paradigms, \model{} exposes how systems balance stability and plasticity under controlled update and preservation conditions. 

\model{} is not intended to be an exhaustive leaderboard.
Our current suite is a compact instantiation of the proposed paradigm specifications, and future work can instantiate the same trial structure in other domains, modalities and real user histories. Nevertheless, the framework demonstrates that agent memory evaluation should move from asking only whether a system answers correctly to asking how it updates, preserves and distorts information over time.

\section*{Limitations}

\model{} is a diagnostic framework rather than an exhaustive benchmark. While our new generated suite demonstrates that the paradigm specifications can be systematically scaled beyond the 56-episode reference suite, the current evaluation still covers only a limited range of domains, languages, user histories and deployment settings. We use a unified reader and structured scoring protocol to enable controlled comparison across systems, and additional analysis suggests that the main findings are not driven by this choice, although native end-to-end deployments may differ. Finally, because part of the dialogue surface form is generated with LLM, the suite may still contain generator-specific biases in wording, scenario realization, or interaction style.

\bibliography{custom}

\newpage
\appendix
\appendix

\section{Implementation and Reproducibility Details}
\label{app:implementation}

\subsection{Code and Data Availability}

We provide a complete reproducibility package in our Github (https://github.com/jq-ding/MemProbe). The archive includes all code used to generate and validate the demonstration suite, the full 56-episode dataset and latent specifications, the runner scripts for all evaluated memory systems, raw model outputs, computed metrics, result tables, and analysis scripts. It also contains detailed implementation notes documenting environment setup, model configurations, system-specific adaptations, excluded systems, and troubleshooting details. In addition, we include instructions for constructing customized \model{} suites, including how to define new fact types, create latent specifications, generate dialogues, validate gold labels, and evaluate memory systems under the same protocol.

\subsection{Environment and Model Configuration}

The main environment used a unified Gemini-based evaluation stack. We used \texttt{gemini-3-flash-preview} as the final reader for all systems and \texttt{gemini-2.5-flash} as the internal LLM for memory extraction, consolidation, graph construction, or memory promotion whenever a system required an LLM component and allowed this configuration. When a system required embeddings, we used the system-supported embedding backend; for Gemini-compatible systems, we used \texttt{gemini-embedding-001}. For systems relying on local sentence-transformer embeddings, we used the default or recommended local embedding model.

\paragraph{Model-selection rationale.}
We use Gemini-3-Flash as the primary reader rather than a GPT-family model because a substantial portion of the demonstration-suite dialogue text was generated with GPT-5.4 assistance, although the latent specifications, gold labels, and validation checks were manually reviewed. Using a non-GPT reader reduces generator and reader coupling in the main evaluation. We also ran cross-model validation with GPT-5.4, Gemini-3-Flash, and Qwen3-32B (App.~\ref{app:cross-model-validation}). The qualitative diagnostic patterns were consistent across models, but Qwen3-32B was weaker on source attribution and conflict-status fields, making it less appropriate as the default reader. We therefore select Gemini-3-Flash as a strong independent reader. For internal memory operations, we use Gemini-2.5-Flash when configurable because it is supported by most evaluated systems, provides stable structured extraction, and avoids the extra runtime and output-budget complications associated with thinking-mode models.

The main goal of this configuration was to make final answering comparable across memory systems. Each system was allowed to use its own memory mechanism, but probe answering was standardized through the same reader and structured scorer.

\subsection{Common Evaluation Protocol}

All incremental memory systems were evaluated under the same session-by-session protocol. For each episode, we created a fresh isolated memory store, fed the episode sessions in chronological order, retrieved the system's memory or context at probe time, and then asked the unified reader to answer the probe using that context.

Formally, for each episode, the evaluation followed:

\begin{enumerate}
    \item Reset or create a fresh memory store.
    \item Ingest sessions sequentially in their original order.
    \item For each probe, retrieve memory/context from the system rather than giving the full episode.
    \item Pass the retrieved memory/context to the unified reader.
    \item Score the structured answer using the same typed-field scorer.
\end{enumerate}

Whenever a memory system exposed a context-only retrieval interface, we used it instead of the system's own final answer generation. This ensures that differences in final answers primarily reflect the retrieved or maintained memory rather than differences in the answer generator. If a system did not expose a clean retrieval interface, we treated it as an end-to-end memory system and documented this separately.

\subsection{Evaluated Systems}

We successfully evaluated six incremental memory systems:

\begin{itemize}
    \item \textbf{Mem0}, representing extraction-based long-term memory.
    \item \textbf{LangMem}, representing framework-level memory extraction and consolidation.
    \item \textbf{Graphiti/Zep}, representing temporal knowledge-graph memory.
    \item \textbf{Cognee}, representing graph-vector hybrid memory.
    \item \textbf{A-MEM}, representing agentic self-organizing memory.
    \item \textbf{MemoryOS}, representing hierarchical memory-OS style memory.
\end{itemize}

We also evaluated retrieval-based diagnostic baselines, including Naive RAG, Time-aware RAG, and Oracle RAG. Naive RAG retrieves top-ranked sessions for each probe. Time-aware RAG additionally annotates retrieved sessions with their original temporal position. Oracle RAG bypasses retrieval and supplies clean non-filler evidence sessions, serving as a diagnostic upper bound rather than a deployable baseline.

\subsection{Systems Considered but Excluded}

We attempted to evaluate several additional memory systems but excluded them due to implementation or protocol incompatibilities.

\begin{itemize}
    \item \textbf{MemGPT}~\citep{memgpt} was excluded because the available implementation required a different version of python/server setup and its tool-based memory mechanism was incompatible with our Gemini-controlled function-calling protocol.
    \item \textbf{Hindsight}~\citep{latimer2025hindsight2020buildingagent} was excluded due to database and system dependency constraints, including PostgreSQL/pgvector and glibc incompatibilities in our environment.
    \item \textbf{Memory-R1}~\citep{memoryr1} was excluded because a runnable implementation was not available at the time of our experiments, and the method is an RL fine-tuning framework rather than an inference-time memory system compatible with our protocol.
    \item \textbf{Memoria}~\citep{sarin2025memoriascalableagenticmemory} was excluded due to dependency conflicts with the unified reader stack and the absence of a clean fixed-transcript ingestion API under our setting.
\end{itemize}

These exclusions do not change the goal of our evaluation, which is to diagnose representative memory mechanisms rather than exhaustively rank all available systems. \model{} is intended to diagnose representative memory mechanisms rather than exhaustively rank all available memory products or research prototypes.

\subsection{Artifacts}

For reproducibility, we release the episode suite, latent specifications, runner scripts, raw prediction reports, metric computation scripts, and the full implementation log. The full implementation log contains exact versions, installation notes, system-specific patches, runtime characteristics, and detailed failure modes for excluded systems.

\section{Demonstration Suite Details}
\label{app:suite_details}

Section~\ref{sec:paradigm-specification} defines \model{} as a set of diagnostic paradigm specifications. This appendix describes one concrete instantiation of these specifications: a 56-episode demonstration suite designed to evaluate stability--plasticity behavior in agent memory systems. The suite is not intended to exhaust all possible memory scenarios. Rather, it serves as a controlled and reproducible testbed for showing how the \model{} paradigms can be instantiated, validated, and used to produce diagnostic behavioral profiles.

\subsection{Suite Overview}
\label{app:suite_overview}

The demonstration suite contains 56 episodes, 1,246 ordered sessions, and 224 probe questions. Each episode instantiates one of the four \model{} paradigms: Interference, Misinformation, Consolidation Strength, and Reconsolidation Window. Each episode contains an initial target memory, a controlled perturbation condition, naturalistic filler interactions, and probes targeting different behavioral readouts.

\begin{table}[ht]
\centering
\small
\resizebox{\linewidth}{!}{
\begin{tabular}{l r}
\toprule
\textbf{Suite property} & \textbf{Value} \\
\midrule
Number of episodes & 56 \\
Number of sessions & 1,246 \\
Number of probes & 224 \\
Paradigms & 4 \\
Domains & Personal, Work, Agentic \\
Split-perturbation episodes & 12 \\
Non-split episodes & 44 \\
Probe types & Current, Previous, Change, Source, Conflict, Temporal \\
\bottomrule
\end{tabular}}
\caption{Summary statistics of the 56-episode \model{} demonstration suite.}
\label{tab:suite_stats}
\end{table}

The suite is balanced across the four diagnostic paradigms, with 14 episodes per paradigm. We also balance update-oriented and preserve-oriented conditions so that simple strategies such as always updating or always preserving cannot solve the suite. In addition, the suite contains both local-evidence episodes, where the relevant perturbation is contained in a single interaction, and split-evidence episodes, where the trigger and accepted new value are distributed across multiple sessions.

The 56-episode suite is a reference instantiation of \model{} rather than the specification itself. Other researchers may instantiate the same paradigms with different domains, languages, interaction styles, or memory systems, as long as the controlled relation among the initial memory, perturbation, expected behavior, and probes is preserved.

\subsection{Domains and Fact Types}
\label{app:domains_fact_types}

To avoid limiting evaluation to lifestyle preferences, we instantiate the paradigms across three broad domains: personal memory, work or productivity memory, and agentic or tool-state memory. This reflects the fact that modern agent memory systems are used not only to remember user preferences, but also to maintain project state, configuration values, tool outputs, and task-specific commitments.

\begin{table}[t]
\centering
\small
\begin{tabular}{p{0.23\linewidth} p{0.65\linewidth}}
\toprule
\textbf{Domain} & \textbf{Example fact types} \\
\midrule
Personal / lifestyle &
Current gym, dietary preference, coffee preference, commute plan, routine preference \\

Work / productivity &
Submission deadline, project owner, team standup time, deployment cadence, review schedule \\

Agentic / tool-state &
Reported port number, active branch, selected database, retry limit, configuration value, tool-reported state \\
\bottomrule
\end{tabular}
\caption{Domains and representative fact types used in the demonstration suite.}
\label{tab:domains}
\end{table}

Each fact type is chosen to support both update and preserve conditions. For example, a project owner can change through a confirmed handoff, but a teammate's temporary involvement should not be treated as an ownership transfer. Similarly, a reported port number may be updated by an authoritative deployment log, while a stale tool output should not be accepted as the current state. These distinctions allow similar surface values to play different roles depending on source, status, and target applicability.

\subsection{Construction Pipeline}
\label{app:construction_pipeline}

The suite was constructed using a staged pipeline that separates experimental design from surface dialogue generation. This separation is important because the main object of evaluation is not the wording of any single episode, but the controlled relationship between initial memory, perturbation, expected behavior, and probes.

The construction pipeline consists of five stages:

\begin{enumerate}
    \item \textbf{Fact-type design.}
    We manually define fact types compatible with one or more \model{} paradigms. Each fact type must support clear initial values, plausible updates, plausible distractors, and unambiguous current-state labels.

    \item \textbf{Value-pair design.}
    For each fact type, we define initial values, candidate new values, distractors, and non-target values. Value pairs are reviewed to avoid ambiguity, trivial templates, and obvious shortcuts.

    \item \textbf{Latent specification.}
    Each episode is first represented as a structured latent specification. The specification fixes the paradigm, condition, expected behavior, canonical values, source type, support level, reactivation status, and required probe types.

    \item \textbf{Dialogue generation.}
    Dialogue sessions are generated from the latent specification. Generation is constrained so that the target memory, perturbation, fillers, and probes remain consistent with the structured design. Some filler interactions were adapted from or inspired by public long-memory benchmark materials, including LongMemEval~\cite{wu2024longmemeval}, LoCoMo~\cite{maharana2024evaluating}, and MemoryArena~\cite{he2026memoryarena}, while the target memory manipulations, perturbations, probes, and gold labels were constructed according to our \model{} latent specifications.

    \item \textbf{Validation and cleanup.}
    We apply rule-based checks, structured gold checks, LLM-judge consistency checks, Full-context reader validation, retrieval-based baselines, and heuristic baselines. Episodes with schema inconsistencies, ambiguous gold labels, or accidental updates are revised.
\end{enumerate}

This pipeline keeps the manipulated experimental variable separate from surface dialogue form. It also makes the suite extensible: new domains or fact types can be added by creating new latent specifications that satisfy the same paradigm constraints.

\subsection{Latent Specifications}
\label{app:latent_specifications}

Each episode is grounded in a latent specification that records the controlled variables of the trial. The latent specification is used both for generation and for evaluation. It prevents the dialogue from being treated as unstructured text and ensures that each episode has a well-defined expected behavior.

Each latent specification contains the following fields:

\begin{itemize}
    \item \textbf{Paradigm and condition}: the \model{} paradigm and the specific condition being instantiated, such as \texttt{authoritative\_update}, \texttt{decoy\_no\_update}, \texttt{assistant\_noise}, \texttt{stale\_tool}, \texttt{high\_support\_weak\_challenge}, or \texttt{reactivated\_weak\_evidence}.

    \item \textbf{Expected behavior}: whether the system should update the target memory or preserve the existing memory.

    \item \textbf{Canonical values}: the initial value, accepted new value if applicable, conflict value, distractor values, and previous value.

    \item \textbf{Source and status}: whether the perturbation comes from the user, assistant, third party, tool, stale tool output, or another entity.

    \item \textbf{Support and reactivation variables}: for consolidation and reconsolidation trials, the support level, evidence strength, and whether the prior memory is explicitly reactivated.

    \item \textbf{Split perturbation metadata}: whether the perturbation evidence is contained in one session or split across multiple sessions.

    \item \textbf{Structured gold}: canonical answers for current value, previous value, change status, source, conflict status, and temporal sequence.
\end{itemize}

\subsection{Dialogue and Perturbation Structure}
\label{app:dialogue_perturbation_structure}

Each episode consists of ordered sessions. Sessions are assigned both a global index and a phase label. The main phases are encoding, filler, perturbation, post-perturbation context, and probing.

Encoding sessions establish the initial memory. Filler sessions introduce natural background interactions and near-miss facts. Confusable sessions mention values that are semantically or lexically similar to the target fact but should not determine the gold answer. Perturbation sessions introduce the critical update, conflict, weak evidence, or reactivation. Post-perturbation sessions test whether systems are overly sensitive to recency or surface similarity.

A standard non-split episode follows the pattern:
\begin{equation}
\small
\begin{aligned}
&\text{Encoding} \rightarrow \text{Filler} \rightarrow \text{Perturbation} \\
&\rightarrow \text{Post-perturbation} \rightarrow \text{Probing}.
\end{aligned}
\label{eq:standard_episode_structure}
\end{equation}

In split-perturbation episodes, the perturbation is intentionally distributed:
\begin{equation}
\small
\begin{aligned}
&\text{Encoding} \rightarrow \text{Filler} \rightarrow P_1 \\
&\rightarrow \text{Interleaving filler} \rightarrow P_2 \\
&\rightarrow \text{Post-perturbation} \rightarrow \text{Probing}.
\end{aligned}
\label{eq:split_episode_structure}
\end{equation}

Here, $P_1$ may contain the source, trigger, or reason for change, while $P_2$ contains the accepted current value or final decision. This design prevents a retrieval system from solving the episode by retrieving a single decisive update sentence. Instead, the system must integrate evidence across multiple sessions.

Split perturbation is included to test distributed evidence integration. In many real interactions, an update is not expressed as a single sentence of the form ``the value changed from $x$ to $y$.'' A user may first indicate that a plan has changed, then later mention the accepted new value. Alternatively, a tool may provide a trigger, while a user later confirms how it should be interpreted.

The split design separates three pieces of information that are often collapsed in simple benchmarks:

\begin{itemize}
    \item \textbf{Change trigger}: evidence that some update or reconsideration occurred.
    \item \textbf{Source or authority}: who or what introduced the relevant evidence.
    \item \textbf{Accepted current value}: the value that should be treated as current.
\end{itemize}

A system that only retrieves $P_1$ may know that a change occurred but fail to identify the new value. A system that only retrieves $P_2$ may identify the new value but miss the trigger or source. A robust memory system should connect these pieces of evidence while preserving the previous state.

\subsection{Structured Gold Labels}
\label{app:structured_gold}

Each probe is paired with structured gold labels. We avoid relying only on free-form textual answers because many correct predictions differ in wording. Structured gold fields specify the intended answer at the level of value, source, status, target applicability, and temporal relation.

For current-value probes, the gold label specifies the accepted current value. For previous-value probes, it specifies the earlier value that should remain historically accessible. For change-detection probes, it specifies whether a confirmed change occurred. For source probes, it specifies who or what introduced the relevant evidence. For conflict-value probes, it specifies the conflicting value, its source, its status, and whether it applies to the target fact. For temporal probes, it specifies the old-to-new direction, trigger, reason, and whether an exact time is required.

For example, a conflict gold label has the following abstract structure:

\begin{verbatim}
{
  "conflict_value": "...",
  "mentioned": true,
  "source": "...",
  "status": "<not_accepted | stale |
              other_entity | hypothetical>",
  "applies_to_target": true or false,
  "correct_current_value": "..."
}
\end{verbatim}

Similarly, a temporal gold label has the following structure:

\begin{verbatim}
{
  "change_direction": "<old_to_new | 
                        no_confirmed_change>",
  "old_value": "...",
  "new_value": "...",
  "trigger": "...",
  "reason": "...",
  "exact_time_required": false
}
\end{verbatim}

Structured labels make scoring more robust to paraphrase and allow partial subscores. For example, a system may identify the conflict value but fail to classify whether it was accepted, stale, hypothetical, or associated with another entity.

\section{Probe Taxonomy and Scoring}
\label{app:probe_scoring}

This section describes the probe taxonomy and scoring procedure used in \model{}. The main text reports aggregate and behavioral metrics, while the details below specify how individual model responses are matched against structured gold labels.

\subsection{Probe Types}
\label{app:probe_types}

\model{} uses targeted probes as behavioral readouts rather than ordinary question-answering items. Each probe type isolates a different aspect of memory behavior.

\paragraph{Current-value probes.}
Current-value probes ask for the currently valid value of the target fact. They evaluate whether the system reaches the appropriate final memory state after the episode, either by accepting a valid update or preserving the original value when no valid update occurred.

\paragraph{Previous-value probes.}
Previous-value probes ask for the earlier value after a change. They evaluate whether historical information remains accessible after an update, rather than being erased or overwritten by the current value.

\paragraph{Change-detection probes.}
Change-detection probes ask whether a real update occurred. They distinguish systems that simply output a plausible value from systems that recognize the update status of a memory.

\paragraph{Source probes.}
Source probes ask who or what introduced a relevant value. They evaluate whether the system preserves provenance information, such as whether a value came from the user, an assistant statement, a tool result, a retrieved document, or another source.

\paragraph{Conflict-value probes.}
Conflict-value probes ask whether a competing value was mentioned, where it came from, and whether it applied to the target fact. They test whether the system distinguishes valid evidence from distractors, stale information, misinformation, or values belonging to another entity.

\paragraph{Temporal probes.}
Temporal probes ask about the update sequence, including the original value, current value, trigger, and reason for change. They evaluate whether the system can reconstruct the temporal structure behind a memory update, not only the final state.

Together, these probes separate memory behaviors that would otherwise be collapsed by final-answer accuracy. A system may recover the current value but lose the previous value, detect that a conflicting value was mentioned but misattribute its source, or identify a change without correctly explaining why the change occurred.

\subsection{Scoring Procedure}
\label{app:scoring_procedure}

We score six probe types: current value, previous value, change detection, source, conflict value, and temporal sequence.

\paragraph{Current-value scoring.}
A current-value response is correct if it identifies the accepted current state of the target fact according to the episode gold label.

\paragraph{Previous-value scoring.}
A previous-value response is correct if it recovers the earlier value after an update. If no update occurred and the probe asks for a previous value, the expected answer is determined by the episode label, such as ``no previous value'' or the original value when the wording asks for the initially established state.

\paragraph{Change-detection scoring.}
A change-detection response is correct if it correctly determines whether a confirmed update occurred. This score evaluates update-status recognition rather than value extraction alone.

\paragraph{Source scoring.}
A source response is correct if it identifies the source associated with the relevant value or update. Acceptable answers may include normalized source categories such as user, assistant, tool, document, third party, stale record, or other entity, depending on the episode label.

\paragraph{Conflict-value scoring.}
Conflict-value probes are decomposed into multiple fields: value identification, source identification, status classification, and target applicability. A response may receive partial credit if it identifies the competing value but misclassifies whether that value was accepted, rejected, stale, hypothetical, unreliable, or non-applicable to the target entity.

\paragraph{Temporal scoring.}
Temporal probes are scored using the old value, current value, update direction, trigger, and reason for change. Exact dates or timestamps are required only when explicitly specified by the gold label. Otherwise, a correct response must capture the update sequence and the event or source that triggered the change.

\subsection{Normalization and Partial Credit}
\label{app:normalization_partial_credit}

Before scoring, we normalize common surface variants that do not change meaning. For example, optional prepositions in time expressions may be ignored, and numeric values may be compared with or without common units when the unit is unambiguous. For sentence-level predictions, we extract candidate values when possible before comparison.

Partial credit is used only for structured probes with multiple fields, such as conflict-value and temporal probes. This prevents formatting differences from being counted as semantic errors while preserving important distinctions among value identification, source attribution, status classification, and temporal reasoning.

\subsection{Execution Errors and Exclusions}
\label{app:execution_errors}

API failures, parsing failures, or execution errors unrelated to episode content are marked separately and excluded from the corresponding accuracy denominators. These exclusions are not treated as model reasoning errors. All exclusions are reported in the validation tables associated with the experiment.

\section{Inter-Axis Correlation}
\label{app:correlation}

\model{} reports results along two decomposition axes: the four cognitive paradigms and the six probe types. A natural question is how much independent signal each axis carries. Treating the six memory systems as samples, we compute pairwise correlations between axes and summarize the off-diagonal entries.

\begin{table}[ht]
\centering
\small
\setlength{\tabcolsep}{4pt}
\begin{tabular}{lrrrr}
\toprule
& \multicolumn{2}{c}{\textbf{Pearson}} & \multicolumn{2}{c}{\textbf{Spearman}} \\
\cmidrule(lr){2-3}\cmidrule(lr){4-5}
\textbf{Axis} & \textbf{mean} & \textbf{min} & \textbf{mean} & \textbf{min} \\
\midrule
4 paradigms   & 0.93 & 0.85 & 0.88 & 0.77 \\
6 probe types & 0.78 & 0.43 & 0.75 & 0.43 \\
\bottomrule
\end{tabular}
\caption{Off-diagonal correlation summary across the two decomposition
axes, computed over the six incremental memory systems.}
\label{tab:axis_correlation}
\end{table}

The paradigm axis is highly collinear: all twelve off-diagonal correlations exceed $0.85$, so a system that is strong on one paradigm tends to be strong on all four. Paradigm-level accuracy therefore largely reflects general competence rather than four separable
capabilities. This is expected, and it is consistent with the role the paradigms play in \model{}: they construct mechanistically distinct controlled conditions under which memory behavior can be
observed, not four independent scoring dimensions.

The probe-type axis is measurably less collinear. Its minimum off-diagonal correlation ($0.43$) is well below the paradigm minimum ($0.85$), indicating that some probe types capture behavior the others do not predict. Temporal fidelity is the clearest case: it correlates only $r=0.64$ with current-value accuracy, so reconstructing the update sequence is substantially less correlated to reporting the correct current value. Source fidelity, by contrast, correlates more strongly
with current value ($r=0.89$), but this reflects a uniform deficit rather than a shared capability: every system scores far lower on source than on current value (Table~\ref{tab:failure}). The independent diagnostic signal in \model{} therefore comes primarily from decomposing correctness by probe type.

\section{Suite Quality Control and Validation}
\label{app:suite_validation}

We validate the suite at three levels: schema consistency, gold-label consistency, and behavioral sanity checks. The goal is to ensure that the episodes are internally consistent, answerable under complete evidence, and not solvable by trivial shortcuts.

\subsection{Schema and Gold-Label Validation}
\label{app:schema_gold_validation}

All episodes are checked for required top-level fields, latent-spec fields, session fields, and metadata fields. Each session has a session index, phase label, session type, dialogue content, and metadata. Split episodes explicitly record the number of perturbation sessions and the identities of perturbation parts.

We use structured gold labels for all probe types and manually audit cases where gold labels and dialogue content disagree. We also normalize common value variants, such as optional prepositions in time expressions and unit-bearing numeric values. This reduces false negatives caused by surface mismatch rather than model error.

\subsection{Probe-Level Validation}
\label{app:probe_validation}

Table~\ref{tab:probe_validation} reports validation accuracy by probe type. High judge and Full-context scores indicate that most gold labels are recoverable under complete evidence. The lower RAG scores on current-value and change-detection probes suggest that the suite selectively stresses retrieval and evidence integration rather than merely increasing surface ambiguity.

\begin{table}[ht]
\centering
\small
\resizebox{\linewidth}{!}{
\begin{tabular}{lccc}
\toprule
\textbf{Probe type} & \textbf{Judge} & \textbf{Full-context} & \textbf{Naive RAG} \\
\midrule
Current value    & 98.2\% & 96.4\% & 71.4\% \\
Previous value   & 100.0\% & 100.0\% & 92.9\% \\
Change detection & 100.0\% & 94.6\% & 65.5\% \\
Temporal         & 86.4\% & 85.7\% & 81.8\% \\
Source           & 97.1\% & 97.1\% & 97.1\% \\
Conflict value   & 100.0\% & 96.4\% & 100.0\% \\
\bottomrule
\end{tabular}}
\caption{
Probe-level validation accuracy for the 56-episode suite. Judge and Full-context scores are high for most probe types, indicating that structured gold labels are largely recoverable when all evidence is visible. Naive RAG shows the largest degradation on current-value and change-detection probes, suggesting that retrieval-based systems struggle most when they must identify the accepted current state and determine whether a real update occurred.
}
\label{tab:probe_validation}
\end{table}

\subsection{Behavioral Sanity Checks}
\label{app:behavioral_sanity_checks}

We run several reference baselines before using the suite to evaluate memory systems. A Full-context reader checks whether the episode is solvable when all evidence is visible. A naive RAG baseline checks whether retrieval and evidence integration are non-trivial. Heuristic baselines such as latest-value, always-update, and always-preserve test whether the suite can be solved by simple shortcuts.

Table~\ref{tab:suite_validation} summarizes the current validation results. The judge and Full-context reader indicate that the episodes are largely readable and internally consistent. The RAG gap and low heuristic scores indicate that the suite is not solved by simple retrieval or recency shortcuts.

\begin{table}[ht]
\centering
\small
\begin{tabular}{l r}
\toprule
\textbf{Baseline} & \textbf{Accuracy} \\
\midrule
LLM judge & 97.8\% \\
Full-context reader & 95.5\% \\
Naive RAG & 81.2\% \\
Latest-value heuristic & 18.8\% \\
FC--RAG gap & 14.3 pp \\
\bottomrule
\end{tabular}
\caption{Validation results for the 56-episode demonstration suite. The judge and Full-context reader indicate that the episodes are largely readable and internally consistent, while the RAG gap and low heuristic score indicate that the suite is not solved by simple retrieval or recency shortcuts.}
\label{tab:suite_validation}
\end{table}

\subsection{Split-Perturbation Validation}
\label{app:split_validation}

The split-perturbation subset provides an additional diagnostic check. Full-context performance remains high on split episodes, indicating that these episodes are still answerable when all evidence is visible. However, naive RAG drops substantially on split episodes, showing that retrieval systems struggle when evidence is distributed across sessions.

\begin{table}[t]
\centering
\small
\resizebox{\linewidth}{!}{
\begin{tabular}{l r r r}
\toprule
\textbf{Subset} & \textbf{Judge} & \textbf{Full-context} & \textbf{Naive RAG} \\
\midrule
Split episodes & 95.8\% & 93.6\% & 56.2\% \\
Non-split episodes & 98.3\% & 96.0\% & 88.0\% \\
\bottomrule
\end{tabular}}
\caption{Split versus non-split validation results. Split perturbations preserve Full-context readability but sharply reduce RAG performance, indicating that they diagnose distributed evidence integration rather than simply making the text ambiguous.}
\label{tab:split_validation}
\end{table}

\subsection{Cross-model validation}
\label{app:cross-model-validation}

To check whether the validation results depend on a single reader model, we repeat the suite validation with three readers: GPT-5.4, Gemini-3-Flash, and Qwen3-32B. All models are evaluated with the same structured JSON scoring protocol. The probes remain natural-language questions, but model responses are normalized into structured fields before scoring. This reduces surface-form bias across models while preserving the semantic requirements of each probe.

Table~\ref{tab:cross_model_overall} summarizes the overall results. All three readers achieve high judge accuracy, indicating that the structured gold labels are largely recoverable across models. Full-context performance is consistently higher than naive RAG, showing that retrieval-based reading introduces a substantial additional bottleneck. Qwen3-32B is weaker overall, especially on source and conflict-status reasoning, but it preserves the same qualitative pattern: Full-context reading outperforms naive RAG.

\begin{table}[t]
\centering
\small
\resizebox{\linewidth}{!}{
\begin{tabular}{lccc}
\toprule
\textbf{Baseline} & \textbf{GPT-5.4} & \textbf{Gemini-3-Flash} & \textbf{Qwen3-32B} \\
\midrule
Judge        & 96.9 & 94.2 & 90.2 \\
Full-context & 92.4 & 88.8 & 78.1 \\
Naive RAG    & 68.8 & 68.3 & 59.4 \\
\midrule
FC--RAG gap  & 23.6 & 20.5 & 18.7 \\
\bottomrule
\end{tabular}}
\caption{
Cross-model validation summary under structured JSON scoring. All three readers show the same qualitative trend: Full-context performance is substantially higher than naive RAG, indicating that the suite exposes retrieval and evidence-integration difficulty beyond Full-context readability.
}
\label{tab:cross_model_overall}
\end{table}

Table~\ref{tab:cross_model_probe_fc} breaks down Full-context performance by probe type. Across models, current-value, previous-value, and change-detection probes are consistently strong, showing that the core factual memory state is readable under full evidence. The largest cross-model differences appear in source and conflict-value probes. Qwen3-32B performs competitively on state-oriented probes but drops sharply on provenance and conflict-status probes, suggesting that source attribution and target-applicability classification are harder than factual value extraction for weaker open-weight readers.

\begin{table}[t]
\centering
\small
\resizebox{\linewidth}{!}{
\begin{tabular}{lccc}
\toprule
\textbf{Probe type} & \textbf{GPT-5.4} & \textbf{Gemini-3-Flash} & \textbf{Qwen3-32B} \\
\midrule
Current value    & 96.4 & 92.9 & 94.6 \\
Previous value   & 100.0 & 100.0 & 96.4 \\
Change detection & 96.4 & 98.2 & 94.6 \\
Source           & 76.5 & 73.5 & 41.2 \\
Conflict value   & 92.9 & 78.6 & 42.9 \\
Temporal         & 86.4 & 77.3 & 72.7 \\
\midrule
Overall          & 92.4 & 88.8 & 78.1 \\
\bottomrule
\end{tabular}}
\caption{
Full-context accuracy by probe type across reader models. State-oriented probes are robust across models, whereas source and conflict-value probes reveal larger differences, especially for Qwen3-32B. This suggests that provenance and conflict-status reasoning are more demanding than extracting current or previous values.
}
\label{tab:cross_model_probe_fc}
\end{table}

Overall, the cross-model results support two conclusions. First, the suite is not specific to the GPT-5.4 reader: Gemini-3-Flash and Qwen3-32B recover the main factual memory states under full context, and Gemini closely matches GPT-5.4 in the overall validation pattern. Second, the split-perturbation effect is robust, retrieval-based reading fails much more severely on split episodes than on non-split episodes, especially for probes that require identifying the accepted current value or detecting that a true update occurred. This validates split perturbation as a diagnostic stressor for distributed evidence integration.

\section{A Second Suite Instantiation}
\label{app:gen40}

\model{} is specified as a construction protocol rather than a fixed dataset. To verify that the protocol is portable rather than tailored to the 56-episode suite, we instantiated a second and independent suite using the same five-stage pipeline (App.~\ref{app:suite_details}) and evaluated the same six memory systems on it.

\subsection{Construction}

The second suite contains 40 episodes, 10 per paradigm, balanced across the same three domains. Two properties make it an independent test of the protocol rather than a re-wording of the original suite.
First, it uses 12 new fact types with no overlap with the original suite, extending the work and agentic domains in particular. Second, dialogue surface form was generated with a different generator model (Gemini-3.1-Pro), so the instantiation does not inherit the wording conventions of the original suite. As before, all controlled variables (latent specifications, perturbations, expected behaviors and gold labels) were human-designed and audited rather than generated.

\subsection{Results}

Because the second suite was generated with a Gemini-family model, we report results under a GPT-5.6 reader, keeping generation and reading in different model families as in the main evaluation.

\begin{table}[ht]
\centering
\small
\resizebox{\linewidth}{!}{
\setlength{\tabcolsep}{3pt}
\begin{tabular}{lrrrrr}
\toprule
\textbf{System} & \textbf{Interf.} & \textbf{Misinf.} & \textbf{Consol.}
& \textbf{Recons.} & \textbf{Overall [95\% CI]} \\
\midrule
A-MEM    & 97.5 & 89.5 & 97.3 & 97.5 & 95.5 [91.7, 98.7] \\
Cognee   & 86.5 & 92.1 & 92.5 & 94.6 & 91.4 [86.3, 96.1] \\
MemoryOS & 84.6 & 76.9 & 68.4 & 80.0 & 77.6 [70.4, 84.3] \\
LangMem  & 82.5 & 70.3 & 86.8 & 63.2 & 75.8 [69.4, 82.6] \\
Graphiti & 75.0 & 71.1 & 66.7 & 83.8 & 74.0 [66.4, 81.8] \\
Mem0     & 45.0 & 34.2 & 43.2 & 48.6 & 42.8 [36.7, 48.4] \\
\bottomrule
\end{tabular}}
\caption{Paradigm-wise and overall accuracy on the second suite.
Confidence intervals are episode-level bootstrap intervals over the
40 episodes.}
\label{tab:gen40_results}
\end{table}

The core diagnostic conclusions replicate. A-MEM and Mem0 again occupy the top and bottom of the ranking, and the source accuracy again falls well below current-value accuracy for every system. Absolute scores are higher than on the original suite, which is consistent with the second suite having shorter episodes and a stronger reader. 

One difference is that MemoryOS ranks third here but fifth on the original suite. The second suite is weighted toward agentic, shorter-horizon fact types, which suits MemoryOS's hierarchical short/mid/long-term organization, whereas the original suite contains longer episodes with more competing-fact interference.

\section{Limitations of the Demonstration Suite}
\label{app:suite_limitations}

The demonstration suite is LLM-assisted and intentionally controlled. This improves interpretability and validation, but does not capture the full diversity of real user logs. The covered domains are representative rather than exhaustive, and the suite does not include all possible uses of long-term agent memory.

The current instantiation is also text-only and English-only. It does not evaluate multimodal memory, multilingual interaction, or memory updates grounded in long-running external environments. 

These limitations motivate larger and noisier instantiations rather than changing the role of \model{}. Future work can instantiate the same paradigm specifications in multilingual, multimodal, user-derived, or application-specific settings while preserving the controlled variables that make stability--plasticity behavior measurable.

\section{License and Artifact Usage}
\label{app:license}

\subsection{License}
We use standard licenses from the community and provide the following links to the licenses for the datasets, codes, and models that we used in this paper:

\vspace{3pt}
\noindent\textbf{Mem0:} \href{https://github.com/mem0ai/mem0/blob/main/LICENSE}{Apache}

\vspace{3pt}
\noindent\textbf{Zep/Graphiti:} \href{https://github.com/getzep/graphiti/blob/main/LICENSE}{Apache}

\vspace{3pt}
\noindent\textbf{LangMem:} \href{https://github.com/langchain-ai/langmem/blob/main/LICENSE}{MIT}

\vspace{3pt}
\noindent\textbf{Cognee:} \href{https://github.com/topoteretes/cognee/blob/main/LICENSE}{Apache}

\vspace{3pt}
\noindent\textbf{A-MEM:} \href{https://github.com/agiresearch/A-mem/blob/main/LICENSE}{MIT}

\vspace{3pt}
\noindent\textbf{MemoryOS:} \href{https://github.com/BAI-LAB/MemoryOS/blob/main/LICENSE}{Apache}

\vspace{3pt}
\noindent\textbf{LoCoMo:} \href{https://github.com/snap-research/locomo/blob/main/LICENSE.txt}{Attribution-NonCommercial 4.0 International
}

\vspace{3pt}
\noindent\textbf{LongMemEval:} \href{https://github.com/xiaowu0162/LongMemEval/blob/main/LICENSE}{MIT}

\vspace{3pt}
\noindent\textbf{MemoryArena:} \href{https://huggingface.co/datasets/ZexueHe/memoryarena}{CC-BY-4.0}

\subsection{Artifact Usage}
The use of existing artifacts is consistent with their intended use in this work.
We will make our code and models publicly accessible and all created artifacts will be only for research purposes and should not be used outside of research contexts.

\subsection{AI Assistants Usage}
We strictly follow the ARR rules in AI Assistants usage and carefully check all the produced artifacts.

\end{document}